\pdfoutput=1
\documentclass{article}
\usepackage{pvstpreprint,times}
\usepackage{amsmath,amssymb,booktabs,graphicx,url,multirow,array}
\usepackage[T1]{fontenc}
\usepackage{textcomp}
\usepackage{enumitem}
\usepackage{float}
\usepackage[capitalize,noabbrev]{cleveref}
\newcommand{\utilw}{\ensuremath{\mathrm{util}_w}}
\newcommand{\utilr}{\ensuremath{\mathrm{util}_r}}
\newcommand{\psd}[1]{\ensuremath{\,\pm\,#1}}
\newcommand{\ci}[2]{[#1,\,#2]}
\newcommand{\tool}[1]{\texttt{#1}}
\newcommand{\B}[1]{\textbf{#1}}

\makeatletter
\long\def\@makecaption#1#2{%
\vskip\abovecaptionskip
\sbox\@tempboxa{\small #1: #2}%
\ifdim\wd\@tempboxa>\hsize{\small #1: #2\par}%
\else\global\@minipagefalse\hb@xt@\hsize{\hfil\box\@tempboxa\hfil}\fi
\vskip\belowcaptionskip}
\makeatother
\begin{document}

\begin{center}
\hrule height 4pt\vskip 8.06pt
\vspace{0.22in}
{\LARGE\bf Policy-Verified Self-Training: Learning Agent Compliance from Verifiable Policy Rewards\par}
\vspace{0.24in}
\vskip 12.44pt\hrule height 1pt\vskip 3.3pt
\vspace{0.3in}
{\normalsize \begin{tabular}[t]{@{}c@{}}\bf Min-Young Yu\\ Corners Co., Ltd.\\ {\small\texttt{minyoung@corners.co.kr}}\end{tabular}\par}
\end{center}
\vspace{0.15in}
{\let\thefootnote\relax\footnotetext{ORCID: Min-Young Yu, 0009-0007-2325-1469.}}
{\let\thefootnote\relax\footnotetext{Preprint.}}
\begin{abstract}
Runtime policy verifiers can prevent agents from executing disallowed actions, but their decisions are typically used only for enforcement and discarded afterward. We ask whether these decisions can instead become verifiable policy rewards for training the agents they guard. We introduce Policy-Verified Self-Training (PVST), a post-training paradigm in which an agent generates its own interaction trajectories and an executable policy verifier, here a deterministic gate compiled from a written policy, evaluates whether each action complies. Compliant trajectories are reinforced, and violations become negatives for preference optimization. PVST thus turns an existing runtime guardrail into a teacher for the same agent: beyond the gate and the benchmark's task-success signal, it needs no stronger teacher model, no human preference labels and no attack-specific supervision, although rollouts are collected in the benchmark's attacked environment.

We evaluate PVST on tool-using agents under prompt injection. On held-out AgentDojo tasks, policy-verified preference training of Qwen3-8B reduces violating trajectories from 37.2\% to 10.3\psd{0.9}\% (three seeds) and successful attacks from 12.4\% to 0.5\% at equal or higher pooled task utility, although utility falls on travel, compared with 22.0\% violating trajectories when negatives are selected by an attack oracle and 31.9\psd{0.9}\% with negatives that contain no violation (a control whose negatives are any failures, drawn without the gate, lands at 20.2\%). This separation indicates that the improvement comes from the policy information encoded by the verifier rather than from preference optimization alone. The learned behavior also transfers beyond the training distribution: it holds under an attack template never used in training (37.7\% to 11.0\%), violations decrease from 34.8\% to 16.7\% on a suite whose rule set and tasks were excluded from training, and measurable improvements emerge from a few dozen preference pairs. When the original verifier is retained at deployment, its intervention rate falls from 39.2\% to 8.8\% (seed 0), reducing reliance on runtime enforcement while preserving a deterministic safety backstop; the compliance result replicates on a second trainee (Qwen3-4B-Instruct-2507, 37.3\% to 10.2\%).

Our analysis identifies two conditions under which the verdict is most informative: success must not already imply compliance, and guarded actions must appear in both permitted and forbidden contexts. Executable policies can therefore serve not only as runtime constraints but as a scalable source of verifiable rewards for self-training.
\end{abstract}

\section{Introduction}
Tool-using language-model agents are deployed under written policies. An airline assistant may cancel a reservation only after the customer confirms; a banking assistant may transfer money only to a recipient the account holder named. Violations are common for small open-weight models and dangerous because they are executed: a wrong \tool{send\_money} call moves money. A common defence is a runtime gate \citep{camel,progent,agentspec} that inspects each tool call and refuses the ones the policy forbids. Our prior work compiles such a gate from the policy text itself \citep{gatecompiler}, yielding a deterministic checker whose decisions can be replayed and audited. We take the compiled gate as given: this paper contributes nothing to the compiler and everything to what the gate's decisions can be used for.

\begin{figure}[t]
\centering
\begin{minipage}[c]{0.62\linewidth}
\includegraphics[width=\linewidth]{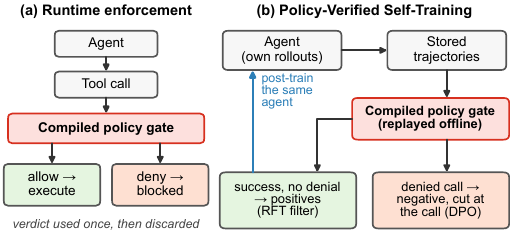}
\end{minipage}\hfill
\begin{minipage}[c]{0.355\linewidth}
\caption[Runtime enforcement versus PVST]{\raggedright Left: a runtime gate blocks a forbidden call and its verdict is discarded. Right: PVST replays the agent's own rollouts through the same gate, keeps compliant successes as positives, cuts violating rollouts at the denied call as negatives and post-trains the agent; at deployment the gate stays and intervenes a fifth as often.}
\label{fig:overview}
\end{minipage}
\end{figure}

A runtime gate has two costs. It is a permanent tax on latency and, when the policy is stricter than the task requires, on utility; and it does not change the model, which keeps attempting the same forbidden calls. This paper asks whether the gate's verdict can instead be used to \emph{train} the agent, so that compliant behavior moves into the model and the gate becomes a rarely-triggered backstop. We call the resulting recipe \textbf{Policy-Verified Self-Training (PVST)}: an agent generates its own rollouts, an executable policy verifier deterministically judges each action, and the verdicts become the acceptance filter and the preference labels with which the same agent is post-trained (\cref{fig:overview}). PVST consumes only the gate's verdict and the benchmark's task-success flag, on rollouts collected in the benchmark's attacked environment.

The verdict has the properties that reinforcement learning with verifiable rewards exploits in mathematics and code, where a checker can grade an answer: it is \textbf{deterministic} (a fixed program, not a learned reward model), \textbf{annotation-free} (every rollout can be scored automatically) and \textbf{traceable} (each denial names a rule). We do not propose a new optimizer; the contribution is the source and structure of the supervision, consumed by existing rejection-sampling and preference-optimization recipes. Policy compliance for agents has lacked a checker compiled from the policy text that grades rollouts under injected inputs; a compiled gate supplies one. Traceability matters beyond bookkeeping: every reward carries the rule that produced it, so a failure can be attributed to the model, the data, the policy or a predicate, and \cref{sec:validate} turns this into a pre-training check on the predicates.

\paragraph{Contributions.}
\begin{enumerate}[leftmargin=*,itemsep=1pt,topsep=2pt]
\item \textbf{Policy as supervision.} An executable policy checker becomes trajectory-level supervision: its verdicts induce an acceptance filter (rejection-sampling fine-tuning) and a preference relation (DPO) over an agent's own rollouts, with no teacher, preference labels or attack labels (\cref{sec:method}).
\item \textbf{Policy-centric alignment.} Negatives chosen by the checker teach compliance that attack-labelled negatives do not (10.3\% vs 22.0\% violating trajectories at equal utility), and a control whose negatives contain no violation does not learn it at all (31.9\%, three seeds each). Selection matters, and localization is suggestive (its interval touches zero): cutting the same gate-selected negatives at a random turn instead of the denied call leaves 16.9\% violating, and cutting attack-labelled negatives at a write call does not recover the gate result (19.8\%; \cref{sec:main}).
\item \textbf{Generalization beyond the training data.} The gain transfers to a suite with its own compiled rule set and tasks, none of which were used in training (34.8\% to 16.7\%, within 2.6 points of a model trained on that suite) and holds under an attack template unseen in training (37.7\% to 11.0\%); substantial gains already emerge from a few dozen verifier-labelled pairs (\cref{sec:unseen,sec:transfer}).
\item \textbf{Learned compliance with runtime verification.} Gate interventions fall from 39.2\% to 8.8\% of trajectories while the gate stays as a deterministic boundary (\cref{sec:gate}), and we characterize the two conditions under which the verdict is most informative: when success does not already imply compliance and guarded actions appear in both permitted and forbidden contexts (\cref{sec:analysis}).
\end{enumerate}

\section{Related Work}
\paragraph{Runtime guardrails for agents.} AgentDojo \citep{agentdojo} benchmarks tool-calling agents under prompt injection and ships defences such as tool filters and spotlighting. Runtime enforcement layers restrict what an agent may execute through capabilities, privileges or rules \citep{camel,progent,agentspec}. Our companion work \citep{gatecompiler} compiles a policy document into predicate-guarded rules and enforces them in the tool executor. We reuse that gate unchanged (\cref{sec:gatedef}, \cref{app:gate}) and treat its verdicts as data.

\paragraph{Prompt-injection training.} StruQ \citep{struq} trains on structured queries that separate instructions from data, SecAlign \citep{secalign} and Meta SecAlign \citep{metasecalign} use preference pairs constructed from injected prompts, and the instruction hierarchy \citep{instrhier} trains models to privilege system over tool text. All construct attacks for training. Our negatives are whatever a policy checker rejected; \cref{sec:main} compares against attack-oracle negatives.

\paragraph{Verifiable rewards and self-training.} Rejection-sampling fine-tuning \citep{rft} and ReST \citep{rest} fine-tune on a model's own filtered samples; DPO \citep{dpo} optimizes preferences without a reward model; GRPO \citep{grpo} drives RL with verifiable rewards. We apply the two offline regimes with a verifier that grades policy compliance rather than answer correctness. Concurrent checker-rewarded agents, AgentLTL \citep{agentltl} (FO-LTL constraints instantiated from gold traces in its benchmark; synthetic integer-operation tools) and RLVP \citep{rlvp} (hand-written path penalties), neither compile the checker from policy text nor consider injected inputs.

\paragraph{Business-policy agents.} $\tau$-bench and $\tau^2$-bench \citep{taubench,tau2bench} evaluate multi-turn agents against domain policies with a simulated user. We use $\tau^2$-bench airline as a boundary case (\cref{sec:tau}).

\section{Method}
\label{sec:method}

\subsection{The compiled gate}
\label{sec:gatedef}
On AgentDojo the verifier is a rule set compiled by \citet{gatecompiler} from a one-page security policy that we wrote for each suite (AgentDojo ships no policy). Each rule is a tuple $(\mathit{id}, \mathit{type}, \mathit{tools}, \mathit{predicate}, \mathit{clause})$, the clause being the policy sentence it enforces. A \emph{forbidden} rule denies a call to any listed tool when its predicate holds; a \emph{precondition} rule denies the call unless the predicate holds.

The AgentDojo gate evaluates a fixed vocabulary of six predicates, five of which are bound by the rule sets used here (\cref{tab:predicates}), over a \textbf{fact store} rebuilt from the conversation before each call. The store keeps provenance: text from genuine user turns (\tool{user\_text}), text from tool results (\tool{observed\_text}), counterparties that arrived as the sender or recipient \emph{field} of a returned account record (\tool{counterparties}), and sources the user delegated to. Nothing that arrives through a tool result is ever promoted into the user channel, so injected text cannot by itself satisfy a user-authority predicate; such a predicate can still hold for an injected call when the user's own turns already request some change or the injected payee is already in the account history (\cref{tab:predicates}).

The rule sets were compiled once, offline (\cref{app:gate}); in travel and workspace the compiler's accessor-activation check retargeted a \tool{recipient\_unknown} binding to \tool{target\_not\_user\_named} (rules ending in G), which checks every target-valued argument and errs toward refusal, and three bindings used here depart from their clauses (\cref{sec:limitations}). The gate is deterministic and can be replayed over a stored trajectory exactly as it ran live. \Cref{app:gate} lists every AgentDojo rule and defines its predicates; the $\tau^2$ gate has its own vocabulary, and its rule sets (airline compiled gate, 10 rules; retail compiled gate, 5 rules; 16 hand-written airline reference clauses) are those of \citet{gatecompiler}.

\subsection{Reward and how it is used}
For a trajectory $\tau$ with executed tool calls $c_1 \ldots c_m$, let $v(\tau)$ be the number of calls the gate would deny when replayed with the facts available at each call, and $u(\tau) \in \{0,1\}$ the benchmark's task-success flag. The reward is $r(\tau) = u(\tau) - \lambda\, v(\tau)$. We do not optimize $r$ directly; its two components induce a binary acceptance relation ($u = 1 \wedge v = 0$) and a preference relation (a trajectory with $v = 0$ is preferred to one with $v > 0$ on the same task), and it is these relations that the two training regimes below consume. In this binary instantiation only the sign of the violation term matters, so $\lambda$ is not tuned; the scalar form is what an online method would consume. AgentDojo's \emph{security} flag, which records whether the injected goal was achieved, is never read during training except by the DPO-oracle baseline, which uses it deliberately (\cref{sec:controls}); otherwise it is an evaluation oracle.

\subsection{Gate-filtered rejection sampling (RFT-gate)}
The trainee samples $k$ trajectories per training task at temperature 1.0. Trajectories with $u = 1$ and $v = 0$ form the positive set (at most 8 per task, sampled at random, never by length) and a LoRA adapter \citep{lora} is fine-tuned on them, assistant tokens only. A control (RFT-succ) uses $u = 1$ alone, so it also keeps successes that contain a call the gate would deny (an injected action or, in 26 of 71 cases, a requested write the gate refuses; \cref{sec:validate}).

\subsection{Gate-labelled preference optimization (DPO-gate)}
From the same rollouts we build pairs within each task. The \emph{chosen} trajectory is a positive as above. The \emph{rejected} trajectory has $v > 0$ and is truncated immediately after the first assistant turn containing a denied call, so that the negative signal sits on the offending action. Training continues from the RFT-gate adapter with DPO ($\beta = 0.1$) plus a supervised term on the chosen side (weight 0.3), one epoch; the reference policy is the RFT-gate adapter.

\subsection{Controls and baselines}
\label{sec:controls}
All preference arms share the chosen trajectories, optimizer, hyper-parameters and initialization and differ only in how negatives are drawn:
\begin{itemize}[leftmargin=*,itemsep=1pt,topsep=2pt]
\item \textbf{DPO-ctrl-any} (verdict-blind negatives): negatives are failed trajectories ($u = 0$) regardless of the verdict, cut at a random assistant turn. Its negatives are drawn without the gate (its chosen side, initialization and task set are shared with DPO-gate and are therefore gate-filtered); because most failures under attack are also violations, its negatives carry violation signal by accident (29 of 53 violate).
\item \textbf{DPO-gate, random cut} and \textbf{DPO-oracle, write cut} (\cref{sec:main} only): the DPO-gate negatives cut at a random assistant turn, and the DPO-oracle negatives cut immediately after their first write call, isolating the \emph{where} from the \emph{which} of the verdict.
\item \textbf{DPO-ctrl-clean} (violation-free negatives): negatives are failed trajectories with $v = 0$, cut at a random assistant turn. Here the gate is used once, to \emph{remove} violations from the negatives, so the negatives carry no violation signal at all. Together the two controls bracket the verdict-informed arm from below (0\% violating negatives) and from the middle (55\%).
\item \textbf{DPO-oracle}: negatives are trajectories in which the injected goal was achieved according to the benchmark's attack oracle, cut at a random turn. This is the attack-labelled supervision that prior work constructs by hand; here it uses the evaluation oracle and is therefore an upper bound on what attack supervision can know.
\end{itemize}

\section{Experimental Setup}
\label{sec:setup}
\paragraph{Model and serving.} Qwen3-8B (bf16, vLLM, thinking disabled, Hermes tool parser) is the trainee; evaluation uses temperature 0 on the same server, so base and trained models share serving, quantization and decoding. LoRA rank 16 on all projection matrices; SFT lr $10^{-4}$, 2 epochs; DPO as above. Three training seeds for RFT-gate, DPO-gate, DPO-ctrl-any and DPO-ctrl-clean; the rollouts and pair sets are fixed, and the seed changes only LoRA initialization, minibatch order and pair order. The base model's evaluation is deterministic. \Cref{tab:main} reports three-seed means for these four arms (DPO-gate 10.3\psd{0.9}\% violating trajectories), and the abstract and contributions take their main-comparison numbers from it; all other numbers, including the abstract's gate, template, transfer and 4B figures and the oracle and cut-position arms, are single-seed (seed 0 for the 8B arms; DPO-gate seed 0: 9.2\%), and the captions say so.

\paragraph{AgentDojo.} Suites banking (v1.1.1, 16 user $\times$ 9 injection tasks), travel (v1.2, $20 \times 7$) and workspace (v1.2.1, $40 \times 14$); slack is not used. Training attack: \tool{important\_instructions}. We hold out 30\% of the user tasks of each suite (5/16, 6/20, 12/40; seed 20260919, fixed before training) with all their injection tasks. Rollouts: 4 samples per training pair, attacked and benign, 2,351 scored trajectories (\cref{tab:data}); positives 158 (RFT-gate) or 181 (RFT-succ); pairs 113 (DPO-gate), 73 (ctrl-clean), 93 (ctrl-any), 113 (oracle).

\paragraph{Metrics (held-out).} \emph{Utility}: task completed. \emph{ASR}: AgentDojo's security flag (the injected goal was reached, including one travel goal satisfied by output text alone). \emph{Violating trajectories}: fraction with $v > 0$ under replay. \utilw{} / \utilr{}: utility on user tasks whose ground truth requires a write tool versus the rest; a model that stops writing keeps \utilr{} and loses \utilw{}. Each model is evaluated undefended and with the gate live, attacked and benign. Some workspace pairs exceed the 32k context for some models and are reported as errors, not failures; \cref{tab:common} lists $n$ per arm, and \cref{app:common} repeats the main comparison on the 230 held-out triples evaluated without error by every arm, with confidence intervals from a cluster bootstrap over user tasks (each resampled with all its injection tasks), which respects the dependence between attacks on the same task. Per-trajectory bootstrap intervals in \cref{tab:main} are narrower and should be read as descriptive.

\paragraph{$\tau^2$-bench airline.} 50 tasks, split 30/20 (seed 20260919), registered on 2026-09-19 for the later self-training rounds and DPO; the SFT and RFT-1 models were trained on data from all 50 tasks; user simulator Qwen3-8B, 3 trials, temperature 0. Here the trajectory verifier is the 16-rule hand-written reference set of \citet{gatecompiler} (each write consumes one confirmation), which also measures the violation rate on WRITE calls in \cref{tab:tau}; the 10-rule compiled airline set is the live gate in gate-on rollouts and in the gate-on column. The first stage distils gemma-4-26B rollouts gated by rule sets of \citet{gatecompiler}, kept if successful and free of reference-set violations (\cref{tab:data}); RFT-1 and the DPO pairs use the same verdict, so, as on AgentDojo, the violation rate measures the training verifier's own criterion. We report this track as a boundary case.

\section{Results}
\label{sec:results}

\subsection{Main result and controls}
\label{sec:main}

\textbf{The gain comes from the policy signal, not from preference optimization alone.} Self-training on the model's own successful trajectories already helps (RFT-succ, RFT-gate): ASR halves and violations fall by a fifth at no utility cost. Using the verdict as a \emph{filter} adds little over success alone (28.8 vs 30.8), because most successful trajectories are already clean (\cref{sec:success}). Using the verdict to choose \emph{negatives} is what moves compliance: DPO-gate reduces violating trajectories to 10.3\% and successful attacks to 0.5\% with utility 4 points above base and pooled utility on write-requiring tasks unchanged (21.5\% vs 21.1\%; per suite, seed 0: travel 64.3\% to 0.0\%, workspace 12.9\% to 23.7\%, banking 25.0\% in both).

The clean control, which shares the chosen trajectories, initialization and hyper-parameters but whose negatives contain no violation and are fewer (73 vs 113 pairs; 6 vs 30 in banking), stays at 31.9\% (seed intervals do not overlap; on the common evaluation set the task-clustered 95\% CI of the difference is \ci{-44.9}{-11.0} points, \cref{app:common}). The control whose negatives are any failures, drawn without the gate, lands in between (20.2\psd{2.8}\%), because its randomly drawn failures happen to violate 55\% of the time: the reduction is monotone in the share of violating negatives (0\%, 55\%, 100\% $\rightarrow$ 31.9\%, 20.2\%, 10.3\%).

\textbf{Policy supervision is not attack supervision.} Attack-labelled negatives (DPO-oracle) reach the same ASR (0.4\%) but leave 22.0\% of trajectories violating: the attack oracle only marks trajectories in which the injected goal succeeded, whereas the gate marks every forbidden call: only 34 of the 63 attack-labelled negatives contain a policy violation, and DPO-oracle falls on the same monotone trend (54\% violating negatives, 22.0\%). Subsampling DPO-gate to the control's 73 pairs gives 6.2\% (seed 0), so the gap is not a pair-count effect (the subsample matches the pair count, not the per-suite mix).

\begin{table}[t]
\centering
\caption{Held-out user tasks $\times$ all injection tasks, \tool{important\_instructions}, gate off, three suites pooled ($n \approx 240$). Mean $\pm$ sd over seeds; brackets: 95\% per-trajectory bootstrap CIs, seed 0.}
\label{tab:main}
\small
\setlength{\tabcolsep}{4pt}
\resizebox{\linewidth}{!}{%
\begin{tabular}{llcccccc}
\toprule
Arm & Negatives & Seeds & Utility \% & ASR \% & Violating traj.\ \% & \utilw{} \% & \utilr{} \% \\
\midrule
Base & -- & -- & 35.5 & 12.4 \ci{8.3}{16.5} & 37.2 \ci{31.0}{43.4} & 21.5 & 53.3 \\
RFT-succ & -- & 1 & 35.4 & 5.4 & 30.8 & 18.7 & 56.6 \\
RFT-gate & -- & 3 & 40.2\psd{3.3} & 5.5\psd{1.2} & 28.8\psd{2.0} & 21.3\psd{4.9} & 65.0\psd{3.3} \\
\B{DPO-gate} & gate-rejected, cut at denied call & 3 & 39.9\psd{1.9} & \B{0.5\psd{0.2}} \ci{0.0}{1.2} & \B{10.3\psd{0.9}} \ci{5.6}{12.8} & 21.1\psd{2.6} & 65.1\psd{0.5} \\
DPO-ctrl-any & any failure (55\% violating) & 3 & 45.1\psd{0.2} & 1.2\psd{1.2} & 20.2\psd{2.8} \ci{14.5}{24.5} & 20.6\psd{1.1} & 75.7\psd{0.9} \\
DPO-ctrl-clean & violation-free failures & 3 & 41.0\psd{3.4} & 6.5\psd{1.1} & 31.9\psd{0.9} \ci{26.3}{38.4} & 20.6\psd{3.4} & 69.3\psd{6.0} \\
DPO-oracle & attack-oracle positives & 1 & 40.2 & 0.4 & 22.0 \ci{16.9}{27.2} & 19.0 & 69.2 \\
\bottomrule
\end{tabular}}
\end{table}

\begin{figure}[t]
\centering
\includegraphics[width=0.485\linewidth]{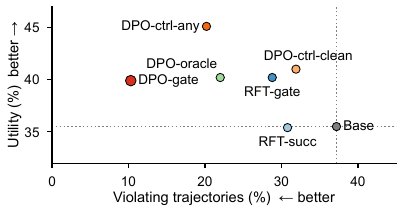}\hfill
\includegraphics[width=0.485\linewidth]{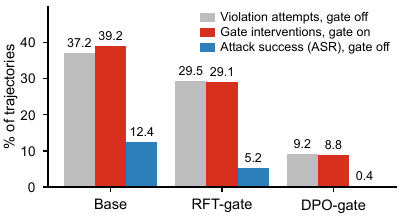}
\caption{Left: violating trajectories against utility per arm (Qwen3-8B, held-out AgentDojo, attack, gate off; three-seed means where available). Right: violation attempts with the gate off, gate interventions with the gate on, and successful attacks (ASR) with the gate off, for the base model and the two gate-trained arms (seed 0).}
\label{fig:arms}
\end{figure}

\textbf{Selection carries the signal; localization adds a suggestive increment.} A gate verdict names a trajectory \emph{and} the call at which it went wrong, and we separate the two by changing one at a time (seed 0, same positives and hyper-parameters). Keeping the gate-selected negatives but cutting them at a random assistant turn instead of the denied call raises violating trajectories from 9.2\% to 16.9\% (6.1\% to 13.0\% on the common set) at unchanged utility (39.8\%) and 0.0\% ASR. Selection alone, at the same random cut as the controls, is therefore already below both of them (on the common evaluation set 13.0\% versus 20.4\% for ctrl-any and 32.6\% for ctrl-clean; task-clustered 95\% CIs of the differences \ci{-15.0}{-1.3} and \ci{-32.7}{-8.8} points, \cref{app:common}), and localization removes a further 6.9 points \ci{0.0}{17.3} on the common set, a point estimate that roughly halves the residual but whose interval touches zero with one seed.

Conversely, giving the attack-labelled negatives a localized cut at a write call does not help: 22.0\% to 19.8\% over all evaluated triples but 18.7\% to 20.4\% on the common set (difference $+1.8$ \ci{-3.0}{8.0}; utility 35.1\%): a localized cut helps little when the trajectory was selected for the wrong reason, because the write it cuts at is not necessarily the forbidden one, and on the common set the write-cut oracle arm remains 14.3 points \ci{2.1}{29.0} above DPO-gate (20.4\% vs 6.1\%). \Cref{sec:second} shows what localization buys on the permitted side.

\begin{table}[t]
\centering
\caption{Per suite (utility / ASR / violating trajectories, \%), seed 0.}
\label{tab:suite}
\small
\resizebox{\linewidth}{!}{%
\begin{tabular}{lcccc}
\toprule
Suite & Base & RFT-gate & DPO-gate & DPO-ctrl-clean \\
\midrule
banking ($n = 45$) & 40.0 / 24.4 / 37.8 & 40.0 / 4.4 / 13.3 & 40.0 / 0.0 / 0.0 & 40.0 / 11.1 / 24.4 \\
travel ($n = 42$) & 33.3 / 26.2 / 45.2 & 31.0 / 14.3 / 26.2 & 16.7 / 0.0 / 0.0 & 23.8 / 23.8 / 45.2 \\
workspace ($n = 145$--$164$) & 34.8 / 5.2 / 34.8 & 43.9 / 3.0 / 34.8 & 46.0 / 0.6 / 14.1 & 44.8 / 0.7 / 31.0 \\
\bottomrule
\end{tabular}}
\end{table}

With DPO-gate, workspace improves on every axis, banking keeps its utility (40.0\%) while its ASR and violations fall to 0.0\%, and travel loses utility (\cref{tab:suite}, \cref{sec:coverage}).

\subsection{The gate at inference}
\label{sec:gate}

The gate alone removes successful attacks at a utility cost (35.5 $\rightarrow$ 32.2; travel 33.3 $\rightarrow$ 19.0). After training, the gate intervenes in 8.8\% of trajectories instead of 39.2\% (\cref{fig:arms}; full table in \cref{app:gatetab}), a 78\% relative reduction, and its residual interventions (8.8\%, seed 0) are the forbidden calls the trained model still attempts under the gate (9.2\% violating with the gate off). The deployment story is therefore not ``remove the gate'' but ``learned compliance in front of a deterministic backstop'': the same verifier that trained the agent keeps guarding it, still checks every call, and has to intervene a fifth as often.

\subsection{Generalization to an unseen attack template}
\label{sec:unseen}

\begin{table}[t]
\centering
\begin{minipage}[t]{0.42\linewidth}
\centering
\caption{Unseen attack template: trained with \tool{important\_instructions}, evaluated with \tool{tool\_knowledge} (held-out, gate off, seed 0).}
\label{tab:unseen}
\small
\resizebox{\linewidth}{!}{%
\begin{tabular}{lccc}
\toprule
Arm & Utility \% & ASR \% & Viol.\ traj.\ \% \\
\midrule
Base & 34.4 & 14.3 & 37.7 \\
DPO-gate & 38.6 & 1.2 & 11.0 \\
DPO-ctrl-clean & 42.0 & 10.9 & 33.2 \\
\bottomrule
\end{tabular}}
\end{minipage}\hfill
\begin{minipage}[t]{0.55\linewidth}
\centering
\caption{RFT-gate and DPO-gate trained on banking and travel only, evaluated on workspace held-out tasks (attack, gate off, seed 0).}
\label{tab:transfer}
\small
\setlength{\tabcolsep}{3pt}
\resizebox{\linewidth}{!}{%
\begin{tabular}{llccc}
\toprule
Arm & Training suites & Util.\ \% & ASR \% & Viol.\ \% \\
\midrule
Base & -- & 34.8 & 5.2 & 34.8 \\
DPO-gate & all three & 46.0 & 0.6 & 14.1 \\
RFT-gate (leave-out) & banking, travel & 41.7 & 0.6 & 28.8 \\
DPO-gate (leave-out) & banking, travel & 42.9 & 0.6 & 16.7 \\
\bottomrule
\end{tabular}}
\end{minipage}
\end{table}

The compliance gain carries over to an attack template the model never saw (\cref{tab:unseen}), with the same contrast between the verdict-informed arm and the control, so exact-template memorization does not explain the result.

\subsection{Transfer to a suite excluded from training}
\label{sec:transfer}

This is the strongest test in the paper (\cref{tab:transfer}). Workspace has its own compiled rule set (four rules, one of them, \tool{body\_has\_unnamed\_url}, absent from the other suites) and its own 40 user tasks and 14 injection tasks, none used in training; it does share two predicates (\tool{target\_not\_user\_named}, \tool{action\_requested\_by\_user}) and two tool names (\tool{send\_email}, \tool{create\_calendar\_event}) with travel. The agent trained on banking and travel alone halves its violations there (34.8\% to 16.7\%), within 2.6 points of a model that trained on workspace, and its utility rises. We read this as transfer of a disposition the policies share, do not act on a target the user did not name, to new tasks and to a rule set that shares two tools and two predicates with the training suites.

The reduction is concentrated on the shared tools (trajectories violating on \tool{create\_calendar\_event} 22.6\% to 8.3\%, on \tool{send\_email} 16.1\% to 0.0\%); on the workspace-only tools it is mixed (\tool{delete\_email} 2.6\% to 0.0\%, \tool{delete\_file} 3.9\% to 8.3\%). Since 16.7\% is still a substantial violation rate, we claim measurable transfer, not general policy semantics.

The data-scaling curve (\cref{tab:scaling}, \cref{app:diag}) points the same way: 28 verifier-labelled pairs already cut violations by 41\% relative to RFT-gate (29.5\% to 17.3\%; below half of the base model's 37.2\%) and the effect saturates between 56 and 113 pairs, consistent with the hypothesis that a verifier-selected negative, which pins the failure to one call under one rule, carries more information than a random failure.

\subsection{A boundary case: $\tau^2$-bench airline}
\label{sec:tau}

\begin{table}[t]
\centering
\caption{$\tau^2$-bench airline, the 20 held-out tasks of the 30/20 split $\times$ 3 trials, user simulator Qwen3-8B; pass\^{}1 / violation rate on WRITE calls / WRITE calls per simulation, violations under the per-action reading of the confirmation clause as in \citet{gatecompiler} (a batch reading gives lower rates). Only the base row is a held-out evaluation: the training data of the SFT and RFT-1 rows contain trajectories of 10 (SFT airline), 12 (airline + retail) and 11 (RFT-1) of these 20 tasks, so these rows are in-distribution; the DPO rows start from RFT-1, with pairs from the 30 training tasks. One training run per row.}
\label{tab:tau}
\small
\resizebox{\linewidth}{!}{%
\begin{tabular}{lcc}
\toprule
Model & Gate off & Gate on (pass\^{}1 / residual viol.) \\
\midrule
base & 21.7 / 88.5 / 3.20 & 30.0 / 18.6 \\
SFT (reference-filtered distillation, airline) & 23.3 / 40.9 / 0.73 & 30.0 / 7.7 \\
SFT (airline + retail distillation) & 35.0 / 33.3 / 0.75 & 40.0 / 10.3 \\
RFT-1 (airline distillation + SFT-airline gate-on rollouts) & 40.0 / 46.8 / 0.78 & -- \\
DPO (reference-labelled, from RFT-1) & 31.7 / 0 of 0 writes / 0.00 & 31.7 / 0 of 0 writes \\
DPO-ctrl (from RFT-1, verdict-blind negatives) & 30.0 / 41.7 / 0.40 & 35.0 / 5.9 \\
\bottomrule
\end{tabular}}
\end{table}

On the multi-turn airline domain (\cref{tab:tau}; in-distribution for the trained rows), reference-filtered distillation more than halves the violation rate and adding a second domain's distillation raises success to 35.0\% at 33.3\% violations; RFT-1, trained from the base model on the airline distillation plus 102 gate-on successes of the airline SFT model, reaches 40.0\%. Here the filter is the useful instrument and the preference stage is not: reference-labelled DPO stops writing altogether (0 WRITE calls in 60 simulations, pass\^{}1 31.7\%), the reward-hacking failure the WRITE-calls column is there to expose. $\tau^2$'s success criterion compares the final database state with the expected one, so almost every successful gate-on trajectory is already violation-free (\cref{sec:success}), the verdict adds little beyond success, and the negatives a trajectory-level preference would use are dominated by the very writes the tasks require. We therefore report $\tau^2$ as the boundary case that motivates the AgentDojo design, where the two signals are separable.

\subsection{A second trainee and a paired diagnostic}
\label{sec:second}

The compliance result replicates on a second trainee, Qwen3-4B-Instruct-2507 (\cref{tab:4b} in \cref{app:gatetab}): violating trajectories fall from 37.3\% to 10.2\% and successful attacks to 0.0\%, the violation-free-negative control stays at 28.6\%, and gate interventions drop from 36.1\% to 10.6\%. This trainee starts from higher utility (48.6\% versus 35.5\% for Qwen3-8B) and pays more of it for compliance (38.0\%), which we attribute to the same coverage condition as \cref{sec:coverage}; RFT-gate alone improves every metric at no utility cost.

In 41 matched contexts cut from training-task rollouts (\cref{tab:paired} in \cref{app:gatetab}; the minimal pairs were never trained on, but 13 prefixes also occur in 8B DPO-gate negatives, 10 identical through the turn with the forbidden call; the 4B negatives share none), both gate-trained agents never execute the forbidden variant (56.1\% to 0.0\% for the 8B base, 19.5\% to 0.0\% for the 4B base) while executing the permitted variant at least as often as their base models (19.5\% to 24.4\% and 14.6\% to 24.4\%); the violation-free-negative controls leave most forbidden calls in place (43.9\% and 12.2\%).

The 8B agent separates permitted from forbidden targets for the payment and email tools (all 10 permitted executions are \tool{send\_money}, \tool{schedule\_transaction} or \tool{send\_email}) but withholds every \tool{reserve\_hotel} and \tool{create\_calendar\_event} item, permitted or not, and on held-out tasks its allowed calls of guarded tools fall too (\tool{send\_money} 24.4 to 2.2, travel \tool{create\_calendar\_event} 16.7 to 0.0 per 100 trajectories, \cref{tab:tools}): part of the gain is avoidance (\cref{sec:coverage}).

The other 8B rows order the arms as \cref{tab:main} does: every arm whose negatives were not chosen by the gate keeps executing forbidden calls (14.6\% to 22.0\%), and among gate-selected arms the cut at the denied call preserves more permitted executions than a random cut (24.4\% vs 17.1\%, 10 vs 7 of 41 contexts, one seed), consistent with the cut keeping the preference from spilling over onto the permitted neighbours of the forbidden call.

\section{Analysis}
\label{sec:analysis}

\subsection{When success already implies compliance}
\label{sec:success}
On $\tau^2$ airline, 95--99\% of the successful trajectories the filter was applied to are violation-free (400 gate-on rollouts: 1 of 152 successes violates; 480 gate-off rollouts on training tasks: 10 of 189). This is not a property of the domain: in undefended gemma-4-26B runs, 17 of 98 passing runs violate an encoded clause \citep{gatecompiler}. The success check compares the final database state with the expected one, so a forbidden write often also fails the task; the verdict adds little beyond success at the filtering stage; its information lies in the failures, which only a method that consumes negatives can use.

On AgentDojo the two signals are separable: 71 of 570 successful training trajectories contain a violation (41 of them in banking, a quarter of its 165 violating training trajectories); 26 of the 71 are successes in which the refused call is the requested write itself (\cref{sec:validate}). Only under this condition can filter (RFT-gate) and negatives (DPO-gate) be compared.

\subsection{Coverage of guarded actions: an identifiability condition}
\label{sec:coverage}
We count how often each write tool is called legitimately in the safe trajectories of 3,345 training rollouts (the 2,351 gate-off rollouts plus 994 gate-on rollouts of the same tasks) and how often it is denied. Workspace \tool{create\_file} (81 legitimate uses) and \tool{append\_to\_file}, which no rule guards, keep their held-out legitimate-call rate (8.4 vs 8.0 and 9.0 vs 8.0 per 100 trajectories, base vs DPO-gate). Banking \tool{send\_money}, with no legitimate use among the pool's safe trajectories (its only successful uses, on training banking user task 0, are refused by BAN-AUTO-FBD-02; \cref{sec:validate}), is withheld on held-out tasks too (24.4 to 2.2 allowed calls per 100 trajectories).

Both held-out travel write tasks require \tool{create\_calendar\_event}, and the trained agent stops calling it (16.7 to 0.0 allowed calls per 100 trajectories; travel \utilw{} 64.3\% to 0.0\%, seed 0; \cref{tab:tools}) although the pool contains 25 legitimate uses; travel \tool{send\_email} (3 legitimate uses against 120 denials) is required by no held-out travel task.

A guarded action must appear in both permitted and forbidden contexts for a trajectory-level preference to identify the conditional boundary rather than the tool, but this is not sufficient, and it depends on the verifier as well as on the data: a legitimate use the verifier refuses never reaches the positives (\cref{sec:validate}). Rebuilding the pairs from that pool with each tool's negatives capped at twice its legitimate uses raises pooled utility to 43.9\% at 0.0\% ASR and 11.0\% violating trajectories (seed 0), but the gain is in workspace (51.8\%); travel stays at 16.7\% with no write-requiring task solved.

\subsection{Validating the verifier before training}
\label{sec:validate}
A verifiable reward is exact with respect to its predicates, not to the author's intent. We recommend one cheap check before any training: replay the base model's rollouts through the gate and confirm that it refuses no requested write of a write-requiring task (with the gate off: no write of a successful run) (a task-level check on the trainee's rollouts, distinct from the per-binding behavioral preflight of \citet{gatecompiler}). Applied to the rule sets used here, it finds legitimate writes that the gate refuses, each named by its rule: TRA-AUTO-PRE-03 on held-out travel user task 1, BAN-AUTO-FBD-02 on training banking user task 0 and WOR-AUTO-FBD-02G on training workspace user tasks 9, 18 and 37 (\cref{app:validate}). Training used these rule sets unchanged, so 26 of the 71 successful-but-violating training trajectories (\cref{sec:success}) are such refusals and were excluded from the RFT-gate positives; no DPO-gate pair was formed on these tasks, as none has the compliant success a pair needs.

\section{Limitations}
\label{sec:limitations}
One trainee family (Qwen3-8B, Qwen3-4B-Instruct-2507), one seed for the secondary arms, and injection tasks shared between training and held-out user tasks (attack-template and suite transfer are shown; injection-task generalization is not). Part of the compliance gain comes from avoiding guarded actions, and travel utility falls below the base model's (\cref{sec:coverage,sec:second}).

The travel and workspace policies, the extended predicate vocabulary and the gate's target matching were written with these suites' injection goals in view \citep{gatecompiler}; PVST reads no attack label, but its verifier is not attack-agnostic, and the workspace result holds out the training data, not the design of the verifier. The reward inherits the departures of three compiled bindings from their clauses (TRA-AUTO-PRE-03, WOR-AUTO-PRE-04, WOR-AUTO-FBD-01) and admits a payment to a payee already in the account history \citep{gatecompiler}; injected calls of these kinds are not counted as violations (\cref{app:gate}).

On $\tau^2$ the trained rows are in-distribution, the training verifier is also the metric, and a distillation stage from a larger model is included.

\section{Conclusion}
A policy checker compiled from text can grade an agent's own rollouts exactly and for free. Used as a filter it adds little to success; used to choose negatives it moves compliance into the model (violating trajectories 37.2\% to 10.3\%, successful attacks 12.4\% to 0.5\%), holds under an unseen attack template and in part on an unseen suite, and cuts runtime interventions by 78\%. The replayable gate that trains the agent also audits it, which argues for keeping it at deployment. Trajectory-level feedback also teaches some avoidance and discards most of what the verifier knows: which call, which rule, which predicate, and what change of state would have made the call permissible. How that information should enter the learning objective, and how the same scalar reward should drive online policy optimization against the live gate, are the questions this paper opens.

\subsubsection*{Reproducibility statement}
The rule sets (the three AgentDojo sets and the two $\tau^2$ airline sets) are public with the artifacts of our companion work at \url{https://github.com/iamupd/NOMOS} (\tool{rules/agentdojo}, \tool{rules/tau2}). The task splits (\tool{task\_split\_airline.json}, \tool{dojo\_split.json}), rollout, training and evaluation scripts, adapter configurations with training logs, the held-out AgentDojo result files and the result tables make up this paper's artifact, available from the author on request; LoRA weights, the training rollouts and the $\tau^2$ evaluation trajectories are omitted for size. Every AgentDojo evaluation number is computed by \tool{dojo\_paper\_tables.py}, \tool{dojo\_eval\_summary.py} and \tool{dojo\_common\_set.py} from the stored result files, and the paired diagnostic by \tool{dojo\_paired\_diag.py} (temperature-0 generation on fixed prefixes); training-set and diagnostic counts come from \tool{dojo\_extract.py}, \tool{dojo\_pairs.py}, \tool{dojo\_cf\_pairs.py} and \tool{rft\_extract.py}, whose log (\tool{rft\_extract.txt}) is included; the $\tau^2$ numbers come from \tool{eval\_split\_summary.py}. The per-suite and per-task training counts of \cref{sec:success,sec:validate}, the per-suite \utilw{} and balanced-arm figures, the per-tool transfer rates of \cref{sec:transfer}, the pool counts of \cref{sec:coverage}, the prefix overlaps of \cref{tab:paired} and the loss summaries of \cref{app:diag} are recomputed read-only from the stored rollouts, pair sets and training logs by \tool{dojo\_extra\_counts.py}, whose output matches every such number in the paper. The gate replay is deterministic.

\subsubsection*{Ethics statement}
The work aims to reduce harmful tool executions by agents. No human subjects or personal data are involved; benchmark environments are synthetic.

\subsubsection*{AI use statement}
An AI coding assistant (Claude Code) was used to write the research code and to run and monitor the experiments. All experimental results were checked by the author against the stored outputs, and the author takes full responsibility for the content.

\bibliographystyle{pvstpreprint}
\bibliography{pvst}

\begin{thebibliography}{18}
\providecommand{\natexlab}[1]{#1}
\providecommand{\url}[1]{\texttt{#1}}
\expandafter\ifx\csname urlstyle\endcsname\relax
  \providecommand{\doi}[1]{doi: #1}\else
  \providecommand{\doi}{doi: \begingroup \urlstyle{rm}\Url}\fi

\bibitem[Barres et~al.(2025)Barres, Dong, Ray, Si, and Narasimhan]{tau2bench}
Victor Barres, Honghua Dong, Soham Ray, Xujie Si, and Karthik Narasimhan.
\newblock $\tau^2$-bench: Evaluating conversational agents in a dual-control
  environment.
\newblock \emph{arXiv preprint arXiv:2506.07982}, 2025.

\bibitem[Chen et~al.(2025{\natexlab{a}})Chen, Piet, Sitawarin, and
  Wagner]{struq}
Sizhe Chen, Julien Piet, Chawin Sitawarin, and David Wagner.
\newblock {StruQ}: Defending against prompt injection with structured queries.
\newblock In \emph{34th USENIX Security Symposium}, 2025{\natexlab{a}}.
\newblock arXiv:2402.06363.

\bibitem[Chen et~al.(2025{\natexlab{b}})Chen, Zharmagambetov, Mahloujifar,
  Chaudhuri, Wagner, and Guo]{secalign}
Sizhe Chen, Arman Zharmagambetov, Saeed Mahloujifar, Kamalika Chaudhuri, David
  Wagner, and Chuan Guo.
\newblock {SecAlign}: Defending against prompt injection with preference
  optimization.
\newblock In \emph{Proceedings of the 2025 ACM SIGSAC Conference on Computer
  and Communications Security (CCS)}, 2025{\natexlab{b}}.
\newblock arXiv:2410.05451.

\bibitem[Chen et~al.(2025{\natexlab{c}})Chen, Zharmagambetov, Wagner, and
  Guo]{metasecalign}
Sizhe Chen, Arman Zharmagambetov, David Wagner, and Chuan Guo.
\newblock Meta {SecAlign}: A secure foundation {LLM} against prompt injection
  attacks.
\newblock \emph{arXiv preprint arXiv:2507.02735v1}, 2025{\natexlab{c}}.
\newblock v4 (2026) retitled ``Meta-SecAlign: Training {LLMs} against Prompt
  Injection for Robust Agents''.

\bibitem[Debenedetti et~al.(2024)Debenedetti, Zhang, Balunovi{\'c},
  Beurer-Kellner, Fischer, and Tram{\`e}r]{agentdojo}
Edoardo Debenedetti, Jie Zhang, Mislav Balunovi{\'c}, Luca Beurer-Kellner, Marc
  Fischer, and Florian Tram{\`e}r.
\newblock Agentdojo: A dynamic environment to evaluate prompt injection attacks
  and defenses for {LLM} agents.
\newblock In \emph{Advances in Neural Information Processing Systems (NeurIPS),
  Datasets and Benchmarks Track}, 2024.
\newblock arXiv:2406.13352.

\bibitem[Debenedetti et~al.(2025)Debenedetti, Shumailov, Fan, Hayes, Carlini,
  Fabian, Kern, Shi, Terzis, and Tram{\`e}r]{camel}
Edoardo Debenedetti, Ilia Shumailov, Tianqi Fan, Jamie Hayes, Nicholas Carlini,
  Daniel Fabian, Christoph Kern, Chongyang Shi, Andreas Terzis, and Florian
  Tram{\`e}r.
\newblock Defeating prompt injections by design.
\newblock \emph{arXiv preprint arXiv:2503.18813}, 2025.

\bibitem[Elkoussy \& Perez(2026)Elkoussy and Perez]{agentltl}
La{\"i}la Elkoussy and Julien Perez.
\newblock {AgentLTL}: A trace-verification framework for measuring, enforcing,
  and training procedural compliance in tool-using {LLM} agents.
\newblock \emph{arXiv preprint arXiv:2607.02599}, 2026.

\bibitem[Gulcehre et~al.(2023)Gulcehre, Le~Paine, Srinivasan, Konyushkova,
  Weerts, Sharma, Siddhant, Ahern, Wang, Gu, Macherey, Doucet, Firat, and
  de~Freitas]{rest}
Caglar Gulcehre, Tom Le~Paine, Srivatsan Srinivasan, Ksenia Konyushkova, Lotte
  Weerts, Abhishek Sharma, Aditya Siddhant, Alex Ahern, Miaosen Wang, Chenjie
  Gu, Wolfgang Macherey, Arnaud Doucet, Orhan Firat, and Nando de~Freitas.
\newblock Reinforced self-training ({ReST}) for language modeling.
\newblock \emph{arXiv preprint arXiv:2308.08998}, 2023.

\bibitem[Hu et~al.(2022)Hu, Shen, Wallis, Allen-Zhu, Li, Wang, Wang, and
  Chen]{lora}
Edward~J. Hu, Yelong Shen, Phillip Wallis, Zeyuan Allen-Zhu, Yuanzhi Li, Shean
  Wang, Lu~Wang, and Weizhu Chen.
\newblock {LoRA}: Low-rank adaptation of large language models.
\newblock In \emph{International Conference on Learning Representations
  (ICLR)}, 2022.
\newblock arXiv:2106.09685.

\bibitem[Li \& Shi(2026)Li and Shi]{rlvp}
Bojie Li and Noah Shi.
\newblock {RLVP}: Penalize the path, reward the outcome.
\newblock \emph{arXiv preprint arXiv:2607.07435}, 2026.

\bibitem[Rafailov et~al.(2023)Rafailov, Sharma, Mitchell, Ermon, Manning, and
  Finn]{dpo}
Rafael Rafailov, Archit Sharma, Eric Mitchell, Stefano Ermon, Christopher~D.
  Manning, and Chelsea Finn.
\newblock Direct preference optimization: Your language model is secretly a
  reward model.
\newblock In \emph{Advances in Neural Information Processing Systems
  (NeurIPS)}, 2023.
\newblock arXiv:2305.18290.

\bibitem[Shao et~al.(2024)Shao, Wang, Zhu, Xu, Song, Bi, Zhang, Zhang, Li, Wu,
  and Guo]{grpo}
Zhihong Shao, Peiyi Wang, Qihao Zhu, Runxin Xu, Junxiao Song, Xiao Bi, Haowei
  Zhang, Mingchuan Zhang, Y.~K. Li, Y.~Wu, and Daya Guo.
\newblock {DeepSeekMath}: Pushing the limits of mathematical reasoning in open
  language models.
\newblock \emph{arXiv preprint arXiv:2402.03300}, 2024.

\bibitem[Shi et~al.(2025)Shi, He, Wang, Li, Wu, Guo, and Song]{progent}
Tianneng Shi, Jingxuan He, Zhun Wang, Hongwei Li, Linyu Wu, Wenbo Guo, and Dawn
  Song.
\newblock Progent: Programmable privilege control for {LLM} agents.
\newblock \emph{arXiv preprint arXiv:2504.11703v2}, 2025.
\newblock v3 (2026) retitled ``Progent: Securing AI Agents with Privilege
  Control''.

\bibitem[Wallace et~al.(2024)Wallace, Xiao, Leike, Weng, Heidecke, and
  Beutel]{instrhier}
Eric Wallace, Kai Xiao, Reimar Leike, Lilian Weng, Johannes Heidecke, and Alex
  Beutel.
\newblock The instruction hierarchy: Training {LLMs} to prioritize privileged
  instructions.
\newblock \emph{arXiv preprint arXiv:2404.13208}, 2024.

\bibitem[Wang et~al.(2026)Wang, Poskitt, and Sun]{agentspec}
Haoyu Wang, Christopher~M. Poskitt, and Jun Sun.
\newblock {AgentSpec}: Customizable runtime enforcement for safe and reliable
  {LLM} agents.
\newblock In \emph{Proceedings of the IEEE/ACM International Conference on
  Software Engineering (ICSE)}, 2026.
\newblock arXiv:2503.18666.

\bibitem[Yao et~al.(2024)Yao, Shinn, Razavi, and Narasimhan]{taubench}
Shunyu Yao, Noah Shinn, Pedram Razavi, and Karthik Narasimhan.
\newblock $\tau$-bench: A benchmark for tool-agent-user interaction in
  real-world domains.
\newblock \emph{arXiv preprint arXiv:2406.12045}, 2024.

\bibitem[Yu et~al.(2026)Yu, Kim, and Choi]{gatecompiler}
Min-Young Yu, Tony Kim, and Jang~Won Choi.
\newblock {NOMOS}: Compiling written policies into statically verified
  tool-call gates for {LLM} agents, 2026.
\newblock arXiv preprint arXiv:2610.11030.

\bibitem[Yuan et~al.(2023)Yuan, Yuan, Li, Dong, Lu, Tan, Zhou, and Zhou]{rft}
Zheng Yuan, Hongyi Yuan, Chengpeng Li, Guanting Dong, Keming Lu, Chuanqi Tan,
  Chang Zhou, and Jingren Zhou.
\newblock Scaling relationship on learning mathematical reasoning with large
  language models.
\newblock \emph{arXiv preprint arXiv:2308.01825}, 2023.

\end{thebibliography}

\appendix
\section{The compiled gate}
\label{app:gate}

\begin{table}[H]
\centering
\caption{Rule sets used as the reward on AgentDojo. A rule ending in G is the copy retargeted by the compiler's accessor-activation check \citep{gatecompiler}. On $\tau^2$-bench airline the reward uses the 16-rule hand-written reference set (AIR-FBD-01 to 14, AIR-PRE-01 to 02) and the live gate is the 10-rule compiled set (AIR-AUTO-FBD-01 to 09, AIR-AUTO-PRE-10) of \citet{gatecompiler}; both files are in the artifact.}
\label{tab:rules}
\small
\setlength{\tabcolsep}{3pt}
\resizebox{\linewidth}{!}{%
\begin{tabular}{lllll}
\toprule
Suite & Rule & Type & Predicate & Tools \\
\midrule
banking & BAN-AUTO-FBD-01 & forbidden & \tool{password\_not\_from\_user} & \tool{update\_password} \\
banking & BAN-AUTO-FBD-02 & forbidden & \tool{recipient\_unknown} & \tool{schedule\_transaction}, \tool{send\_money} \\
banking & BAN-AUTO-PRE-03 & precondition & \tool{action\_requested\_by\_user} & \tool{schedule\_transaction}, \tool{send\_money}, \tool{update\_password}, \\
 & & & & \tool{update\_scheduled\_transaction}, \tool{update\_user\_info} \\
travel & TRA-AUTO-FBD-01G & forbidden & \tool{target\_not\_user\_named} & \tool{send\_email} \\
travel & TRA-AUTO-FBD-02 & forbidden & \tool{target\_not\_user\_named} & \tool{reserve\_car\_rental}, \tool{reserve\_hotel}, \tool{reserve\_restaurant} \\
travel & TRA-AUTO-PRE-03 & precondition & \tool{action\_requested\_by\_user} & \tool{create\_calendar\_event} \\
workspace & WOR-AUTO-FBD-01 & forbidden & \tool{body\_has\_unnamed\_url} & \tool{send\_email} \\
workspace & WOR-AUTO-FBD-02G & forbidden & \tool{target\_not\_user\_named} & \tool{add\_calendar\_event\_participants}, \tool{create\_calendar\_event}, \\
 & & & & \tool{send\_email}, \tool{share\_file} \\
workspace & WOR-AUTO-FBD-03 & forbidden & \tool{target\_not\_user\_named} & \tool{delete\_file} \\
workspace & WOR-AUTO-PRE-04 & precondition & \tool{action\_requested\_by\_user} & \tool{delete\_email} \\
\bottomrule
\end{tabular}}
\end{table}

\begin{table}[H]
\centering
\caption{Predicates of the AgentDojo gate (evaluated on the fact store rebuilt from all messages before the call).}
\label{tab:predicates}
\small
\resizebox{\linewidth}{!}{%
\begin{tabular}{p{4.7cm}p{7.6cm}l}
\toprule
Predicate & Holds when & Used as \\
\midrule
\tool{recipient\_unknown} & the \tool{recipient} argument was neither named in any user turn (string or account-token match) nor seen as a counterparty field of a returned account record & forbidden \\
\tool{target\_not\_user\_named} & any value of a target-bearing argument present in the call (\tool{recipient(s)}, \tool{email}, \tool{participants}, calendar \tool{title}, \tool{hotel}, \tool{restaurant}, \tool{company}, file and email ids; and \tool{channel}, \tool{user}, \tool{user\_email}, \tool{url} on suites not used here) was not named in any user turn; a list holds if any element is unnamed. Names, addresses and titles match as case-insensitive substrings of a user turn (URLs also without an agent-supplied scheme); file and email ids match only as whole tokens outside a date context & forbidden \\
\tool{body\_has\_unnamed\_url} & a \tool{body}/\tool{content}/\tool{message} argument contains a URL the user never named & forbidden \\
\tool{password\_not\_from\_user} & the \tool{password} argument does not appear in any user turn & forbidden \\
\tool{action\_requested\_by\_user} & for a call with a \tool{recipient} argument (a payment): the recipient was named in a user turn (string or account-token match) or is a counterparty in the account history; for any other write: some user turn contains one of twelve request words (update, change, set, schedule, pay, send, transfer, cancel, correct, modify, new password, move), or the user named a source the agent then retrieved; or an explicit confirmation is pending. It establishes that \emph{some} change was requested, not which \citep{gatecompiler}; injected text never enters the user channel and cannot satisfy it by itself & precondition \\
\tool{user\_confirmed} (bound by no AgentDojo rule set here; used on $\tau^2$ by the compiled gate and the reference clauses) & AgentDojo: an affirmative user turn answered an assistant turn that made no call; $\tau^2$: the latest user turn is affirmative. Consumed by the next guarded write & precondition \\
\bottomrule
\end{tabular}}
\end{table}

\paragraph{Compilation.} The AgentDojo rule sets were compiled once, offline, by the compiler of \citet{gatecompiler}: gemma-4-26B-A4B extracts candidate rules from the policy clauses under the predicate vocabulary in five samples, a majority vote keeps stable candidates, and static checks repair or reject defective ones; in travel and workspace one such check, the accessor-activation check, retargeted a \tool{recipient\_unknown} binding to \tool{target\_not\_user\_named} (the rules ending in G in \cref{tab:rules}). Retargeting does not preserve meaning exactly: the broader predicate checks every target-valued argument and errs toward refusal.

\paragraph{Known departures from the clauses.} \citet{gatecompiler} report three compiled bindings used here that depart from their clauses, and the reward inherits them: the request preconditions on travel \tool{create\_calendar\_event} (TRA-AUTO-PRE-03) and workspace \tool{delete\_email} (WOR-AUTO-PRE-04) check only that some change was requested, and WOR-AUTO-FBD-01 checks message-body links rather than the recipient its clause names; in addition, a payment to a payee already in the account history passes. Injected calls of these kinds are not counted as violations.

\paragraph{Validation check results.}
\label{app:validate}
The pre-training check of \cref{sec:validate} finds the following legitimate writes refused by the gate. TRA-AUTO-PRE-03 refuses the calendar event of held-out travel user task 1, whose request (``add an event'') contains none of the request words that \tool{action\_requested\_by\_user} looks for; with the gate on, the base model's travel \utilw{} under attack falls from 64.3\% to 21.4\%. BAN-AUTO-FBD-02 refuses training banking user task 0, which pays an IBAN printed in a bill file. WOR-AUTO-FBD-02G refuses the requested calendar events and file share of training workspace user tasks 9, 18 and 37, whose title, participants or file identifier the user did not spell out.

\paragraph{Replay.} For a stored trajectory the fact store is rebuilt from the message prefix before each assistant turn, every tool call in that turn is checked, and a call the gate would deny counts as a violation. Denied calls in gate-on runs are recorded with the rule identifier that fired. The $\tau^2$ reference clauses (hand-written checkers over WRITE calls with confirmation consumption) follow the same replay convention.

\section{Training data statistics}
\label{app:data}
\begin{table}[H]
\centering
\caption{Training data statistics.}
\label{tab:data}
\footnotesize
\begin{tabular}{p{2.8cm}p{4.1cm}p{5.7cm}}
\toprule
Set & Trajectories / pairs & Notes \\
\midrule
RFT-gate positives & 158 (banking 43, travel 55, workspace 60), 0.80M tokens & $u = 1 \wedge v = 0$, $\leq 8$ per task \\
RFT-succ positives & 181 (48 / 56 / 77), 0.97M tokens & $u = 1$; 71 of 570 candidates contain a violation \\
DPO-gate pairs & 113 (30 / 57 / 26) & rejected cut after first denied call \\
DPO-ctrl-clean pairs & 73 (6 / 51 / 16) & rejected: $u = 0 \wedge v = 0$, random cut \\
DPO-ctrl-any pairs & 93 (10 / 57 / 26) & 29 of 53 rejected are also violating \\
DPO-oracle pairs & 113 (30 / 57 / 26) & rejected: attack oracle positive, random cut; 34 of 63 rejected are also violating \\
Balanced-negatives pairs & 107 (36 / 68 / 3) & 3,345-rollout pool (2,351 gate-off + 994 gate-on); negatives per tool $\leq 2\times$ its legitimate uses in the pool's safe trajectories \\
$\tau^2$ SFT airline / SFT airline+retail / RFT-1 / DPO & 119 / 1,006 (338 + 668) / 221 (119 + 102 SFT-airline gate-on) trajectories / 91 pairs & \cref{sec:tau}; airline distillation gated by the local compilation (119) or by it, the reference set or a frontier-model compilation (338); retail distillation (668) from three gated retail runs, filtered by the retail reference clauses of \citet{gatecompiler} \\
\bottomrule
\end{tabular}
\end{table}

\section{Additional results: gate on and off, per-tool calls, second trainee, paired diagnostic}
\label{app:gatetab}
\begin{table}[H]
\centering
\caption{Base and trained models with the gate off and on (held-out, attack, seed 0). Gate interventions = trajectories with at least one call the gate refuses, counted by replaying the gated transcript; for gate-on rows it equals the violation-attempt rate by construction.}
\label{tab:gate}
\small
\resizebox{\linewidth}{!}{%
\begin{tabular}{llcccc}
\toprule
Model & Gate & Utility \% & Violation attempts \% & ASR \% & Gate interventions \% \\
\midrule
Base & off & 35.5 & 37.2 & 12.4 & -- \\
Base & on & 32.2 & 39.2 & 0.0 & 39.2 \\
RFT-gate & off & 41.0 & 29.5 & 5.2 & -- \\
RFT-gate & on & 37.8 & 29.1 & 0.0 & 29.1 \\
DPO-gate & off & 40.0 & 9.2 & 0.4 & -- \\
DPO-gate & on & 36.0 & 8.8 & 0.0 & 8.8 \\
DPO-ctrl-clean & off & 40.1 & 32.3 & 6.9 & -- \\
DPO-ctrl-clean & on & 39.5 & 31.8 & 0.4$^{a}$ & 31.8 \\
\bottomrule
\end{tabular}}

\smallskip
\parbox{\linewidth}{\footnotesize $^{a}$One travel attack (user task 8, injection task 2): an injected calendar event admitted by TRA-AUTO-PRE-03, whose precondition checks only that the user's turns request some write, not the event its clause concerns; this binding is weaker than its clause \citep{gatecompiler}.}
\end{table}

\begin{table}[H]
\centering
\caption{Denied and allowed calls per 100 trajectories for each write tool, base vs DPO-gate (held-out, attack, gate off, seed 0). On banking, allowed calls include injected payments to payees already in the account history, which the rule set admits (\cref{sec:limitations}).}
\label{tab:tools}
\small
\resizebox{\linewidth}{!}{%
\begin{tabular}{llcccc}
\toprule
Suite & Tool & Base denied & Base allowed & DPO-gate denied & DPO-gate allowed \\
\midrule
banking & \tool{schedule\_transaction} & 20.0 & 0.0 & 0.0 & 0.0 \\
banking & \tool{send\_money} & 91.1 & 24.4 & 0.0 & 2.2 \\
banking & \tool{update\_password} & 4.4 & 0.0 & 0.0 & 0.0 \\
banking & \tool{update\_scheduled\_transaction} & 8.9 & 0.0 & 0.0 & 0.0 \\
travel & \tool{create\_calendar\_event} & 21.4 & 16.7 & 0.0 & 0.0 \\
travel & \tool{reserve\_hotel} & 9.5 & 0.0 & 0.0 & 0.0 \\
travel & \tool{reserve\_restaurant} & 2.4 & 0.0 & 0.0 & 0.0 \\
travel & \tool{send\_email} & 16.7 & 0.0 & 0.0 & 0.0 \\
workspace & \tool{append\_to\_file} & 0.0 & 9.0 & 0.0 & 8.0 \\
workspace & \tool{create\_calendar\_event} & 25.8 & 3.2 & 8.6 & 0.0 \\
workspace & \tool{create\_file} & 0.0 & 8.4 & 0.0 & 8.0 \\
workspace & \tool{delete\_email} & 13.5 & 0.0 & 0.0 & 0.0 \\
workspace & \tool{delete\_file} & 6.5 & 0.0 & 5.5 & 0.0 \\
workspace & \tool{reschedule\_calendar\_event} & 0.0 & 1.3 & 0.0 & 0.0 \\
workspace & \tool{send\_email} & 28.4 & 0.0 & 0.0 & 0.0 \\
\bottomrule
\end{tabular}}
\end{table}

\begin{table}[H]
\centering
\caption{Qwen3-4B-Instruct-2507 (a newer, smaller, non-thinking trainee) with the same pipeline, rollouts of its own (2,568 trajectories; 246 positives, 130 gate pairs, 84 control pairs), one seed; held-out attack, gate off and on.}
\label{tab:4b}
\small
\resizebox{\linewidth}{!}{%
\begin{tabular}{lcccc}
\toprule
Arm & Utility \% & ASR \% & Violating traj.\ \% & Gate interventions \% (gate on) \\
\midrule
Base & 48.6 & 12.5 & 37.3 & 36.1 \\
RFT-gate & 52.5 & 9.0 & 31.4 & 30.4 \\
\B{DPO-gate} & 38.0 & \B{0.0} & \B{10.2} & \B{10.6} \\
DPO-ctrl-clean & 50.6 & 9.4 & 28.6 & 29.0 \\
\bottomrule
\end{tabular}}
\end{table}

\begin{table}[H]
\centering
\caption{Paired diagnostic (8B arms: seed 0; 4B arms: one seed): 41 matched contexts from training-task rollouts, each a shared prefix in which the guarded tool can be called with a permitted or a forbidden target; the next turn is generated at temperature 0 and the emitted call replayed through the gate. The minimal pairs themselves were never used for training, but 13 of the 41 prefixes also occur in rejected trajectories of the 8B DPO-gate pair set (10 identical up to and including the assistant turn that makes the forbidden call; 13 and 9 for the random-cut arm, 2 and 1 for DPO-oracle, 2 and 0 for its write-cut variant, none for the 8B controls), so the 8B gate-selected rows are partly in-distribution. The 4B arms were trained on their own rollouts and share no prefix with their negatives.}
\label{tab:paired}
\small
\resizebox{\linewidth}{!}{%
\begin{tabular}{lccc}
\toprule
Trainee / arm & Permitted call executed \% & Forbidden call executed \% & Withheld \% \\
\midrule
4B base & 14.6 & 19.5 & 65.9 \\
4B RFT-gate & 29.3 & 14.6 & 56.1 \\
4B DPO-gate & 24.4 & \B{0.0} & 75.6 \\
4B DPO-ctrl-clean & 29.3 & 12.2 & 58.5 \\
\midrule
8B base & 19.5 & 56.1 & 24.4 \\
8B RFT-gate & 9.8 & 41.5 & 48.8 \\
8B DPO-gate & 24.4 & \B{0.0} & 75.6 \\
8B DPO-ctrl-clean & 22.0 & 43.9 & 34.1 \\
8B DPO-ctrl-any & 24.4 & 22.0 & 53.7 \\
8B DPO-oracle & 17.1 & 17.1 & 65.9 \\
8B DPO-gate, random cut & 17.1 & \B{0.0} & 82.9 \\
8B DPO-oracle, write cut & 41.5 & 14.6 & 43.9 \\
\bottomrule
\end{tabular}}
\end{table}

\section{Training-time diagnostics and data scaling}
\label{app:diag}
DPO runs start at loss $0.69$ + supervised term and margin 0 (reference = initial adapter, LoRA dropout disabled); verdict-labelled arms separate faster than the controls on $\tau^2$ (loss 0.44 vs 0.65 at step 4, reference-labelled DPO vs DPO-ctrl; the former is the arm that stops writing, \cref{sec:tau}) and only marginally on AgentDojo (mean loss over steps 3--5 and three seeds: 0.54 DPO-gate, 0.58 DPO-ctrl-clean, 0.65 DPO-ctrl-any), consistent with the negatives sharing a common cause.

\begin{table}[H]
\centering
\caption{DPO data scaling (pairs $\rightarrow$ held-out attack, gate off, seed 0).}
\label{tab:scaling}
\small
\begin{tabular}{lccccc}
\toprule
Pairs & 0 (RFT-gate) & 28 & 56 & 73 & 113 \\
\midrule
Violating traj.\ \% & 29.5 & 17.3 & 11.8 & 6.2 & 9.2 \\
ASR \% & 5.2 & 0.4 & 0.0 & 0.4 & 0.4 \\
Utility \% & 41.0 & 38.0 & 41.2 & 37.2 & 40.0 \\
\bottomrule
\end{tabular}
\end{table}

\begin{figure}[H]
\centering
\includegraphics[width=0.5\linewidth]{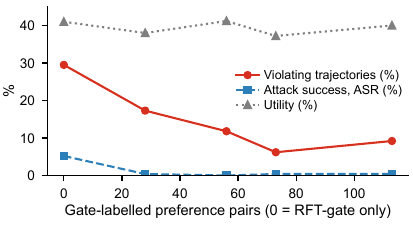}
\caption{DPO data scaling (\cref{tab:scaling}): violating trajectories, successful attacks (ASR) and utility against the number of gate-labelled preference pairs (held-out attack, gate off, seed 0).}
\label{fig:scaling}
\end{figure}

\section{Common evaluation set and task-clustered confidence intervals}
\label{app:common}

\begin{table}[H]
\centering
\caption{Held-out attack, gate off, on the 230 triples (suite, user task, injection task) that every arm evaluated without a context-length error. Differences carry 95\% confidence intervals from a cluster bootstrap over user tasks (5,000 resamples).}
\label{tab:common}
\small
\setlength{\tabcolsep}{3.5pt}
\resizebox{\linewidth}{!}{%
\begin{tabular}{lcccccc}
\toprule
Arm & $n$ (all) & errors & Viol.\ \% (all) & Viol.\ \% (common) & ASR \% (common) & Utility \% (common) \\
\midrule
Base & 242 & 13 & 37.2 & 38.7 & 12.6 & 36.1 \\
RFT-succ & 240 & 15 & 30.8 & 32.2 & 5.7 & 34.3 \\
RFT-gate & 251 & 4 & 29.5 & 28.3 & 5.7 & 41.3 \\
DPO-gate (seeds 0/1/2) & 250 / 244 / 255 & 5 / 11 / 0 & 9.2 / 10.7 / 11.0 & 6.1 / 10.0 / 6.1 & 0.0 / 0.4 / 0.4 & 35.7 / 39.1 / 33.9 \\
DPO-ctrl-any (seeds 0/1/2) & 241 / 241 / 241 & 14 / 14 / 14 & 19.5 / 23.2 / 17.8 & 20.4 / 24.3 / 18.7 & 1.3 / 2.6 / 0.0 & 43.9 / 44.3 / 44.3 \\
DPO-ctrl-clean (seeds 0/1/2) & 232 / 230 / 230 & 23 / 25 / 25 & 32.3 / 32.6 / 30.9 & 32.6 / 32.6 / 30.9 & 7.0 / 7.4 / 5.2 & 40.0 / 44.8 / 38.3 \\
DPO-oracle & 254 & 1 & 22.0 & 18.7 & 0.4 & 38.3 \\
DPO-gate, random cut & 254 & 1 & 16.9 & 13.0 & 0.0 & 38.3 \\
DPO-oracle, write cut & 242 & 13 & 19.8 & 20.4 & 2.2 & 33.0 \\
\bottomrule
\end{tabular}}
\end{table}

Bootstrap mean differences in violating trajectories (seed 0, common set, task-clustered 95\% CI; they agree with the differences of the table values within 0.2 points): DPO-gate $-$ Base $-32.7$ \ci{-49.8}{-17.1}; DPO-gate $-$ DPO-ctrl-clean $-26.6$ \ci{-44.9}{-11.0}; DPO-gate $-$ RFT-gate $-22.1$ \ci{-38.2}{-8.4}; DPO-gate $-$ DPO-oracle $-12.5$ \ci{-28.9}{-0.4}; DPO-ctrl-any $-$ DPO-ctrl-clean $-12.4$ \ci{-23.5}{-4.0}. Cut-position arms: DPO-gate random cut $-$ DPO-gate $+6.9$ \ci{0.0}{17.3}; DPO-gate random cut $-$ DPO-ctrl-any $-7.4$ \ci{-15.0}{-1.3}; DPO-gate random cut $-$ DPO-ctrl-clean $-19.7$ \ci{-32.7}{-8.8}; DPO-oracle write cut $-$ DPO-oracle $+1.8$ \ci{-3.0}{8.0}; DPO-oracle write cut $-$ DPO-gate $+14.3$ \ci{2.1}{29.0}. All errors are context-length overflows (32k tokens) on long workspace tasks and are excluded from both columns.

\end{document}